\documentclass[letterpaper, 10 pt, journal, twoside]{IEEEtran}
\ifCLASSINFOpdf
\else
\fi
\let\labelindent\relax
\usepackage{amsfonts}       

\usepackage{amsthm}
\usepackage{mathtools}      
\usepackage{amssymb}        
\usepackage{filecontents}
\usepackage{graphicx}       
\usepackage{marginnote}     
\usepackage{marvosym}       
\usepackage{overpic}        
\usepackage{tabularx}
\usepackage{cite}
\usepackage{color}
\usepackage[dvipsnames]{xcolor}
\usepackage[lined,linesnumbered,noend, ruled]{algorithm2e}
\usepackage{epstopdf}
\usepackage[inline]{enumitem}
\usepackage[normalem]{ulem}

\usepackage{lipsum}
\usepackage{subfig}     
\usepackage{wrapfig}    
\usepackage[font=footnotesize,labelfont=bf]{caption}    
\usepackage{microtype}  
\usepackage{comment}
\usepackage[group-separator={,}]{siunitx}   
\usepackage{titlesec}   

\usepackage{etoolbox}   
\makeatletter
\patchcmd{\@algocf@start}
  {-1.5em}
  {0pt}
  {}{}
\makeatother

\definecolor{darkblue}{rgb}{0.0, 0.0, 0.55}
\makeatletter
\let\NAT@parse\undefined
\makeatother
\usepackage[bookmarks=true, hidelinks]{hyperref}    
\hypersetup{
     colorlinks   = true,
     linkcolor= darkblue,
     citecolor    = darkblue,
     urlcolor=black
}

\theoremstyle{definition}

\theoremstyle{remark}

\SetKwProg{Fn}{Function}{}{}
\SetKwComment{Comment}{$\triangleright$\ }{}

\newcommand{\orc}{\textsc{ORC}\xspace}
\newcommand{\mcts}{\textsc{MCTS}\xspace}
\newcommand{\dipn}{\textsc{DIPN}\xspace}
\newcommand{\gn}{\textsc{GN}\xspace}
\newcommand{\dqn}{\textsc{DQN}\xspace}
\newcommand{\uct}{\textsc{UCT}\xspace}

\def\gcvpg{gc-\textsc{VPG}\xspace}
\def\gopg{go-\textsc{PGN}\xspace}
\def\ours{\textsc{VFT}\xspace}

\DeclareMathOperator*{\argmax}{arg\,max}

\titlespacing\section{0pt}{8pt plus 2pt minus 2pt}{4pt plus 2pt minus 2pt}
\titlespacing\subsection{0pt}{6pt plus 2pt minus 2pt}{3pt plus 2pt minus 2pt}
\begin{document}
%
\title{Visual Foresight Trees for Object Retrieval from Clutter with Nonprehensile Rearrangement}
%
%
%

\author{Baichuan Huang, Shuai D. Han, Jingjin Yu, and Abdeslam Boularias%
\thanks{Manuscript received: May, 5, 2021; Revised July, 19, 2021; Accepted October, 14, 2021.}
\thanks{This paper was recommended for publication by Editor Markus Vincze upon evaluation of the Associate Editor and Reviewers' comments.
This work was supported in part by NSF awards IIS-1845888, IIS-1734492, IIS-1846043, CCF-1934924, IIS-1734492, IIS-1846043, and IIS-2132972.} 
\thanks{B. Huang, S. D. Han, J. Yu, and A. Boularias 
are with the Department of Computer Science, 
Rutgers, the State University of New Jersey, Piscataway, NJ, USA. 
        {\tt\footnotesize \{baichuan.huang, shuai.han, jingjin.yu, 
abdeslam.boularias\}@rutgers.edu}}%
\thanks{Digital Object Identifier (DOI): see top of this page.}
\vspace{-5pt}
}
%
%

\markboth{IEEE Robotics and Automation Letters. Preprint Version. Accepted October, 2021}
{Huang \MakeLowercase{\textit{et al.}}: Visual Foresight Trees for Object Retrieval from Clutter with Nonprehensile Rearrangement} 

%



\maketitle
\begin{abstract}
This paper considers the problem of retrieving an object from many
tightly packed objects using a combination of robotic pushing and 
grasping actions. 
Object retrieval in dense clutter is an important skill for robots
to operate in households and everyday environments effectively.
The proposed solution, Visual Foresight Trees (\ours), intelligently rearranges the clutter surrounding a target object so that it can be 
grasped easily.
Rearrangement with nested nonprehensile actions is challenging as it 
requires predicting complex object interactions in a combinatorially 
large configuration space of multiple objects. 
We first show that a deep neural network can be trained to accurately predict the poses 
of the packed objects when the robot pushes one of them. The predictive 
network provides visual foresight and is used in a tree search as a 
state transition function in the space of scene images. 
The tree search returns a sequence of consecutive push actions yielding 
the best arrangement of the clutter for grasping the target object. 
Experiments in simulation and using a real robot and objects show that the 
proposed approach outperforms model-free techniques as well as model-based 
myopic methods both in terms of success rates and the number of executed actions, on several challenging tasks.

A video introducing \ours, with robot experiments, is accessible at \href{https://youtu.be/7cL-hmgvyec}{\texttt{\textcolor{blue}{https://youtu.be/7cL-hmgvyec}}}. The full source code is available at
\href{https://github.com/arc-l/vft}{\texttt{\textcolor{blue}{https://github.com/arc-l/vft}}}.
\end{abstract}

\begin{IEEEkeywords}
Deep Learning in Grasping and Manipulation, Learning from Experience, Visual Learning
\end{IEEEkeywords}

%
\IEEEpeerreviewmaketitle

\section{Introduction}\label{sec:intro}
%
%
%
%
\IEEEPARstart{I}{n} many application domains, robots are tasked with retrieving objects that are surrounded by multiple tightly packed objects. 
To enable the grasping of target object(s), a robot needs to re-arrange the scene to create sufficient clearance before attempting a grasp.
Scene rearrangement can be achieved through nested sequential push actions, each moving multiple objects simultaneously.
In this paper, we address the problem 
of finding the minimum number of push actions to create a scene where the target object can be grasped and retrieved. 

To solve the object retrieval problem, the robot must imagine how the scene would 
look like after any given sequence of pushing actions, and select the shortest 
sequence that leads to a state where the target object can be grasped. 
The huge combinatorial search space makes this problem computationally challenging, 
hence the need for efficient planning algorithms, as well as fast predictive models 
that can return the predicted future states in a few milliseconds. 
Moreover, objects in clutter typically have unknown physical properties such as mass 
and friction coefficients. While it is possible to utilize off-the-shelf physics engines
to simulate contacts and collisions of rigid objects in clutter, simulation is highly sensitive to the accuracy of the provided mechanical parameters. To overcome the 
problem of manually specifying these parameters, and to enable full autonomy of the 
robot, most recent works on object manipulation utilize machine learning techniques 
to train predictive models from data~\cite{Hafner2020Dream,DBLP:journals/corr/abs-1812-00568,8207585}. The predictive models take the state of the robot's environment a control action as inputs and predict the state after applying the control action.
\begin{figure}[t]
    \centering
    \begin{minipage}{.4\linewidth}
        \subfloat[Hardware setup]{\includegraphics[width = \linewidth, trim = 10 110 30 48, clip]{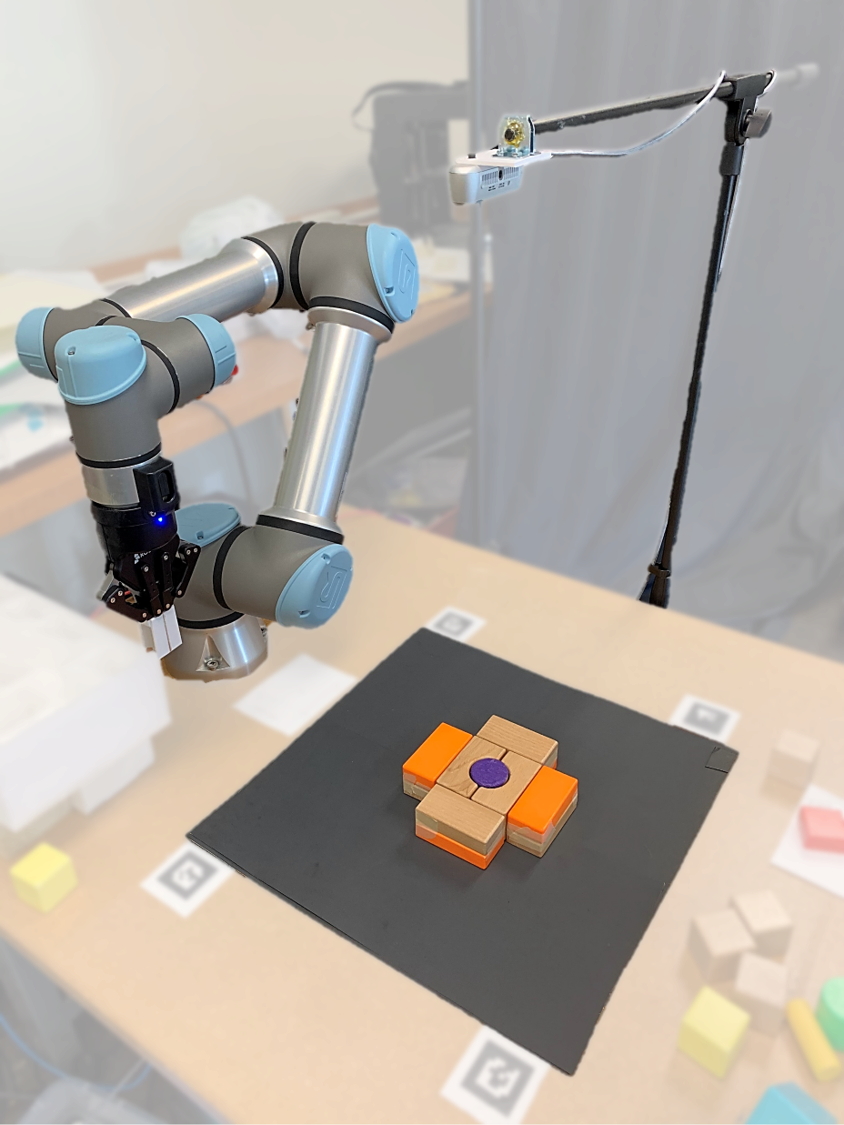}\label{fig:intro-setup}}
    \end{minipage}
    \begin{minipage}{.28\linewidth}
        \subfloat[First push]{\includegraphics[width = \linewidth, trim = 0 0 0 0, clip]{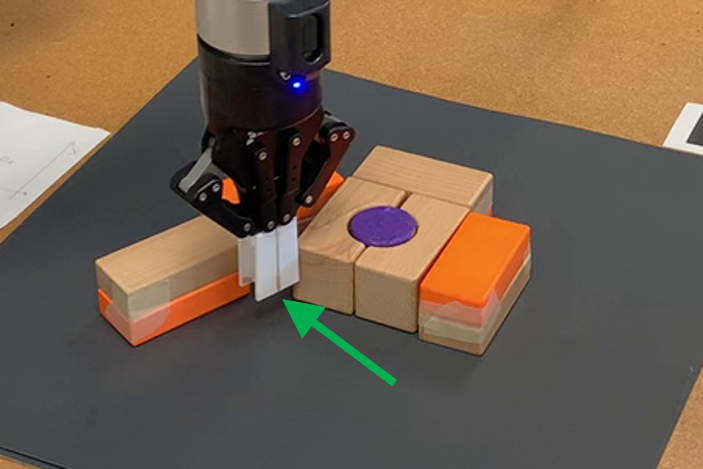}\label{fig:intro-push}}
        
        \subfloat[Second push]{\includegraphics[width = \linewidth, trim = 0 0 0 0, clip]{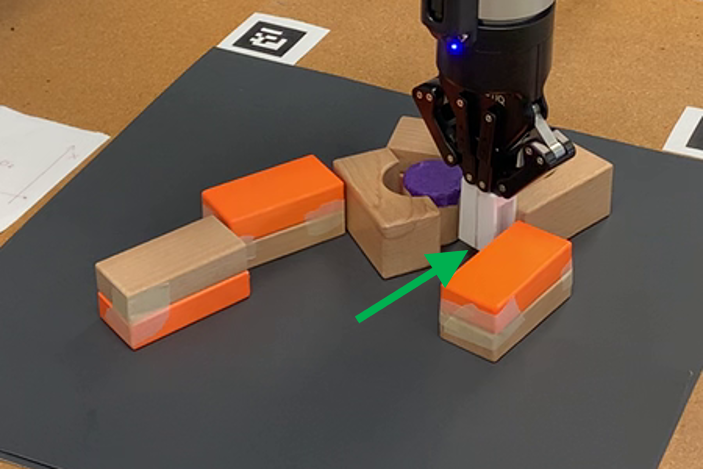}\label{fig:intro-push}} 
    \end{minipage} 
    \begin{minipage}{.28\linewidth}
        \subfloat[Third push]{\includegraphics[width = \linewidth, trim = 0 0 0 0, clip]{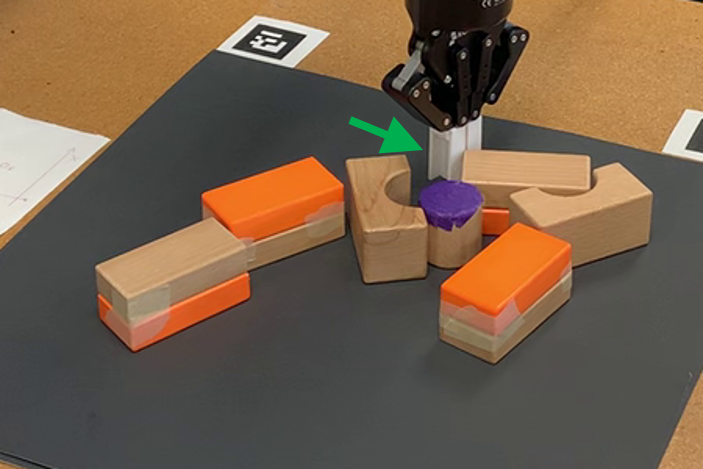}\label{fig:intro-push}}
        
        \subfloat[Grasp]{\includegraphics[width = \linewidth, trim = 0 0 0 0, clip]{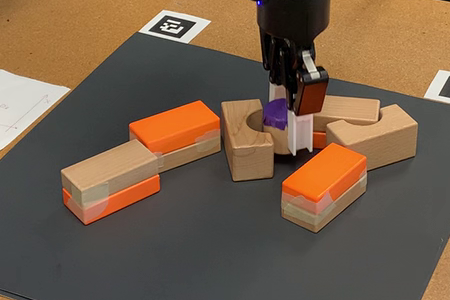}\label{fig:intro-grasp}}
    \end{minipage} 

    \caption{\label{fig:intro}
    (a) The hardware setup for object retrieval in a clutter includes 
    a Universal Robots UR-5e manipulator 
    with a Robotiq 2F-85 two-finger gripper, 
    and an Intel RealSense D435 RGB-D camera. 
    The objects are placed in a square workspace. 
    (b)(c)(d) Three push actions (shown with green arrows) 
    are used to create space 
    accessing the target (purple) object. 
    The push directions are toward top-left, top-right, 
    and bottom-right, respectively. 
    (e) The target object is successfully grasped and retrieved. 
    }
    \vspace{-3mm}
\end{figure}

In this work, we propose to employ {\it visual foresight trees} (VFT) to address the computational and modeling challenges related to the object retrieval problem. 
A key building block of VFT is a Convolutional Neural Networks (CNN) extending DIPN~\cite{huang2020dipn}, capable of predicting multi-step push outcomes involving multiple objects.
A second CNN evaluates the graspability of the target object in predicted future images. 
A Monte Carlo Tree Search utilizes the two CNNs to obtain the shortest sequence of pushing actions that lead to an arrangement where the target can be grasped. 

To our knowledge, the proposed technique is the first model-based learning solution to the object retrieval problem.
Extensive experiments on the real robot and objects are shown in
Fig.~\ref{fig:intro} demonstrate that the proposed approach succeeds in retrieving target objects with manipulation sequences that are shorter than model-free reinforcement learning techniques and a limited-horizon planning technique. 

%

\section{Related works}\label{sec:related}
\noindent {\bf Grasping.} Robotic grasping methods are generally categorized in two main categories: {\it analytical} and {\it data-driven}~\cite{10.1109/TRO.2013.2289018}. 
Analytical approaches rely on precise 3D and mechanical models of objects to simulate {\it force-closure} or {\it form-closure} grasps~\cite{grasping,liang2019pointnetgpd,doi:10.1177/0278364912442972}. Since material properties, such as mass and friction coefficients, are generally difficult to measure, most recent techniques have shifted toward learning directly grasp success probabilities from data. 
Most data-driven methods focused on isolated objects~\cite{DBLP:conf/iros/BoulariasKP11,DBLP:conf/iccv/MousavianEF19, gabellieri2020grasp, lu2020multifingered}. 
Learning to grasp in cluttered scenes was explored in recent works~\cite{DBLP:conf/aaai/BoulariasBS14,mahler2017dexnet,kalashnikov2018qtopt}. 
More recent learning techniques were adapted to grasp objects clutter. For example, a hierarchy of supervisors was used for learning to grasp objects in clutter from demonstrations~\cite{7743488}. CNNs, such as Dex-net 4.0~\cite{mahler2019learning}, were trained to detect grasp 6D poses in point clouds~\cite{DBLP:journals/corr/PasGSP17}.  
A composition of a suction cup and a gripper in~\cite{deng2019deep} was shown to produce more stable grasps learned with CNNs.
A randomized physics-based motion planning technique for grasping in cluttered and uncertain environments was also presented in~\cite{8207585}. A large-scale benchmark for general object grasping was introduced in~\cite{fang2020graspnet}. 
{\it Sim-to-real} transfer was adopted in~\cite{DBLP:journals/corr/abs-1812-07252} for data-efficient learning of robotic grasps.
\vspace{1mm}

\noindent {\bf Object Singulation.} 
A closely related problem is the singulation of individual items~\cite{6224575},
i.e., isolating an item to facilitate its retrieval, 
typically achieved through a combination of pushing and grasping actions.
In contrast with the present work, pushing in singulation is typically performed \emph{model-free} by using reactive policies without explicitly reasoning about future states~\cite{10.1007/978-3-030-28619-4_32}.
Singulation does not generally require long-horizon reasoning.
For instance, linear push policies were learned in~\cite{8560406} to increase grasp access for robot bin picking, by using model-free reinforcement learning. The tasks considered in~\cite{8560406} can be solved through single pushing actions because of the lower density of clutter compared to the tasks considered in our work.
While the focus of the present work is on sequential pushing actions, a new 6-DOF grasping method~\cite{murali20206} was devised to create clearance for an object by picking and placing obstacles away.
The 6-DOF grasping was also combined with a push policy~\cite{tanglearning}. Results reported in ~\cite{huang2020dipn,tanglearning} show, however, this combination~\cite{tanglearning} is less efficient on the same tasks than the one-step reasoning method~\cite{huang2020dipn} that serves as one of the baselines in our experiments.
%


\noindent {\bf Rearrangement Planning.}
Object retrieval in clutter is closely related to rearrangement planning. 
The approach recently presented in~\cite{DBLP:journals/corr/abs-1912-07024} also uses a Monte Carlo tree search, but the objectives of rearrangement tasks are different from ours. 
In object retrieval, we focus on finding the minimum number of pre-grasp pushing actions that lead to grasping a single target object. This objective requires highly accurate predictions of future poses of individual objects in clutter. 

\noindent {\bf Object Retrieval.}
Several other works also addressed the problem of retrieving a target object from clutter. Some of these works focus on online planning for object search under partial observability without learning~\cite{8793494}.
Other related works learn only the quality of pushing and grasping actions\cite{DBLP:journals/corr/abs-1903-01588}, without visual foresight, which is necessary for tightly packed clutter. Similarly, scene exploration and object search were learned using model-free reinforcement learning, based on active and interactive perception~\cite{DBLP:journals/corr/abs-1911-07482}, and teacher-aided exploration~\cite{kurenkov2020visuomotor}. A planning approach with a human operator guiding a robot to reach for a target object in clutter was presented in~\cite{DBLP:journals/corr/abs-1904-03748}. In contrast to these approaches, ours is fully autonomous. The work presented in~\cite{xu2021efficient} is most related to ours, with a similar robotic setup and objects. However, it is based on deep Q-learning, which is model-free and does not predict future states. We show in Section~\ref{sec:experiments} that our model-based technique significantly outperforms the one from~\cite{xu2021efficient} on the same tasks considered in~\cite{xu2021efficient} as well as on more challenging ones. 


\section{Preliminaries}\label{sec:problem}
\subsection{Problem Statement}
%
The Object Retrieval from Clutter (\orc) challenge asks a robot manipulator to retrieve a target object from a set 
of objects densely packed together. The objects may have different shapes, sizes, 
and colors. 
Objects other than the target object are unknown a priori.  
Focusing on a mostly planar setup, the following assumptions are made:
\begin{enumerate*}
\item The hardware setup (Fig.~\ref{fig:intro-setup}) contains a manipulator, a 
planar workspace with a uniform background color, and a camera on top of the 
workspace. 
\item The objects are rigid and are amenable to the gripper's prehensile 
and non-prehensile capabilities, limited to straight-line planar push actions and top-down grasp actions.
%
\item The objects are confined to the workspace without overlapping. As a result, 
the objects are visible to the camera. 
\item The target object, to be retrieved, is visually distinguishable from the others. 
\end{enumerate*}
Under these assumptions, the \emph{objective} is to retrieve only the target 
object, while minimizing the number of pushing/grasping actions that are used. Each grasp or push is considered as one atomic action. While a mostly planar setup is assumed in our experiments, the proposed data-driven solution is general and can be applied to arbitrary object shapes and arrangements. In the experiments, we mainly work with woodblocks; we also evaluate the proposed approach on novel objects such as soapboxes, which are challenging as their widths are 
close to the maximum distances between the gripper's fingers.


\begin{figure*}[ht!]
    \centering
    \includegraphics[width = 0.98\linewidth]{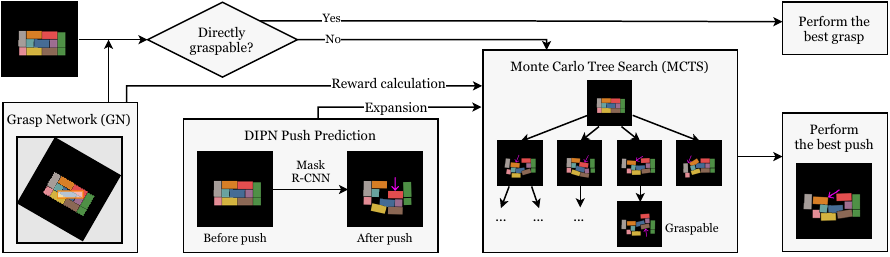}\label{fig:pipeline}
    \caption{\label{fig:pipeline}
        Overview of the proposed technique for object retrieval from clutter with nonprehensile rearrangement. The problem is iteratively solved by observing the environment at each time step, taking the current state as input, and returning the best action. It is repeated until the object is retrieved.
    }
    \vspace{-2mm}
\end{figure*}

\subsection{Manipulation Motion Primitives}
\label{sec:primitives}
Similar to studies closely related to the \orc challenge, e.g.,~\cite{zeng2018learning, huang2020dipn, xu2021efficient}, we employ a set of pre-defined and parameterized 
pushing/grasping manipulation primitives. The decision-making problem then entails 
the search for the optimal order and parameters of these primitives.
A grasp action $a^\text{grasp} = (x, y, \theta)$ is defined as 
a top-down overhead grasp motion at image pixel location $(x, y)$, 
with the end-effector rotated along with the world $z$-axis by $\theta$ degrees. 
In our implementation, a grasp center $(x, y)$ can be any pixel in a 
down-sampled $224 \times 224$ image of the planar scene, while rotation angle 
$\theta$ can be one of $16$ values evenly distributed between $0$ and $2\pi$. 
To perform a complete grasp action, the manipulator moves the open gripper 
above the specified location, then moves the gripper downwards until a contact 
with the target object is detected, closes the fingers, and transfers the 
grasped object outside of the workspace. 

When objects are densely packed, the target object is generally not directly 
graspable due to collisions between the gripper and surrounding objects. 
When this happens, non-prehensile push actions can be used to create 
opportunities for grasping. For a push action $a^\text{push} = (x_0, y_0, x_1, y_1)$, 
the gripper performs a quasi-static horizontal motion. Here, $(x_0, y_0)$ and 
$(x_1, y_1)$ are the start and end location of the gripper center, 
respectively. The gripper's orientation is fixed along the motion direction during 
a push maneuver.

\section{Overview of the Proposed Approach}\label{sec:outline}

When objects are tightly packed, the robot needs to carefully select an appropriate 
sequence of pushes that create a sufficient volume of empty space around the target 
object before attempting to grasp it. 
In this work, we are interested in challenging scenarios where multiple push 
actions may be necessary to de-clutter the surroundings of the target, and where the 
location, direction, and duration of each push action should be carefully optimized 
to minimize the total number of actions. 
Collisions among multiple objects often occur while pushing a single object, further
complicating the matter. 
To address the challenge, we propose a solution that uses a neural network 
to forecast the outcome of a sequence of push actions in the future, and estimates 
the probability of succeeding in grasping the target object in the resulting scene. 
The optimal push sequence is selected based on the forecasts.

A high-level description of the proposed solution pipeline is depicted in Fig.~\ref{fig:pipeline}. 
At the start of a planning iteration, an RGB-D image of the scene is taken, and the objects are detected and classified as {\it unknown clutter} or {\it target object}.
With the target object located, a second network called Grasp Network (\gn) predicts 
the probability of grasping the target. \gn is a Deep Q-Network (\dqn) \cite{mnih2015human}
adopted from prior works~\cite{zeng2018learning, huang2020dipn} for \orc. It 
takes the image input, and outputs the estimated grasp success probability for 
each grasp action. The target object is considered directly graspable if the maximum estimated grasp success probability is larger than a threshold. The robot executes the corresponding optimal grasp action; otherwise, push actions must be performed to create space for grasping. 

When push actions are needed, the next action is selected using Monte-Carlo 
Tree Search (\mcts). In our implementation, which we call the Visual Foresight 
Tree (\ours), each search state corresponds 
to an image observation of the workspace. Given a push action and a state, \ours 
uses the Deep Interaction Prediction Network (\dipn)~\cite{huang2020dipn} as the 
state transition function. Here, \dipn is a network that predicts the motions 
of multiple objects and generates a synthetic image corresponding to the scene 
after the imagined push. \ours uses \gn to obtain a reward value for each search node and detect 
whether the search terminates. Both \dipn and \gn are trained offline on different objects.

\section{Visual Foresight Trees}\label{sec:method}

This section discusses the three main components of \ours: \gn, \dipn, and Monte-Carlo Tree Search (\mcts).

\subsection{Grasp Network}
The Grasp Network (\gn), adapted from \cite{huang2020dipn}, takes the image $s_t$ as input, and outputs a pixel-wise reward prediction $R(s_t) = [R(s_t, a^1),\dots,R(s_t,a^n)]$ for grasps $a^1,\dots,a^n$. 
The output is a 2D map with the same size as the input image, and where each point contains the predicted reward of performing a grasp at the corresponding input pixel.
Table $R(s_t)$ is a one channel image with the same size as input image $s_t$ ($224\times224$ in our experiments), and a value $R(s_t, a^i)$ represents the 
expected reward of the grasp at the corresponding action. 
To train GN, we set the reward to be $1$ for grasps where the robot successfully picks up only the target object, and $0$ otherwise. 
GN is the reward estimator for states in VFT (in Section~\ref{subsec:vft}).


A grasp action $a^\text{grasp} = (x, y, \theta)$ specifies the grasp location and the end-effector angle. 
\gn is trained while keeping the orientation 
of the end-effector fixed relative to the support surface, while randomly varying the poses of the objects. Therefore, \gn assumes that the grasps are aligned to the principal axis of the input image. 
To compute 
reward $R$ for grasps with $\theta \neq 0$, 
the input image is rotated by $\theta$ before passing it to \gn. 
As a result, for each input image, \gn generates $16$ different grasp $R$ reward tables. 



The training process of the \gn used in this work is based on previous works~\cite{zeng2018learning, huang2020dipn} but differs in terms of objectives, which requires a significant modification, explained in the following. 
The objective in previous works is to grasp all the objects; the goal of \orc 
is to retrieve a specific target among a large number of obstacles.
We noticed from our experiments that if \gn is trained to grasp all the objects, then a greedy policy will be learned, and it will always select the most accessible object to grasp. In contrast, all other objects that can also be directly grasped are ignored because they have low predicted rewards. This causes the problem that \gn cannot correctly predict the grasp success rate of a specific target object.
One straightforward adaption to this new objective is only to give reward when the grasp center is inside the target object, which is the approach that was followed in~\cite{xu2021efficient}. 
However, we found that we can achieve a higher sample efficiency by providing a reward for successfully grasping any object.
The proposed training approach is similar in spirit to Hindsight Experience Replay (HER)~\cite{NIPS2017_453fadbd}. 
To balance between exploration and exploitation, grasp actions are randomly sampled from 
$P(s, a^\text{grasp}) \propto b R(s, a^\text{grasp})^{b-1}$ where $b$ is set to $3/2$ in the experiments.


After training, \gn can be used for selecting grasping actions in new scenes.
Since the network returns 
reward $R$ for all possible grasps, and not only for
the target object, the first post-processing step consists in selecting a 
small set of grasps that overlap with the target object. This is achieved by computing 
the overlap between the surface of the target object and the projected footprint 
of the robotic hand, and keeping only grasps that maximize the overlap. 
Then, grasps with the highest predicted values obtained from the trained 
network are ranked, and the best choice without incurring collisions 
is selected for execution. 

\subsection{Push Prediction Network}
\dipn~\cite{huang2020dipn} is a network that takes an RGB-D image, 2D masks of objects, center positions of objects, and a vector of the starting and endpoints of a push action. It outputs predicted translations and rotations for each passed object. 
The predicted poses of objects are then used to create a synthetic image.  
Effectively, \dipn imagines what happens to the clutter if the robot executes 
a certain push.

\begin{figure}[ht!]
    \centering
    \includegraphics[width = \linewidth]{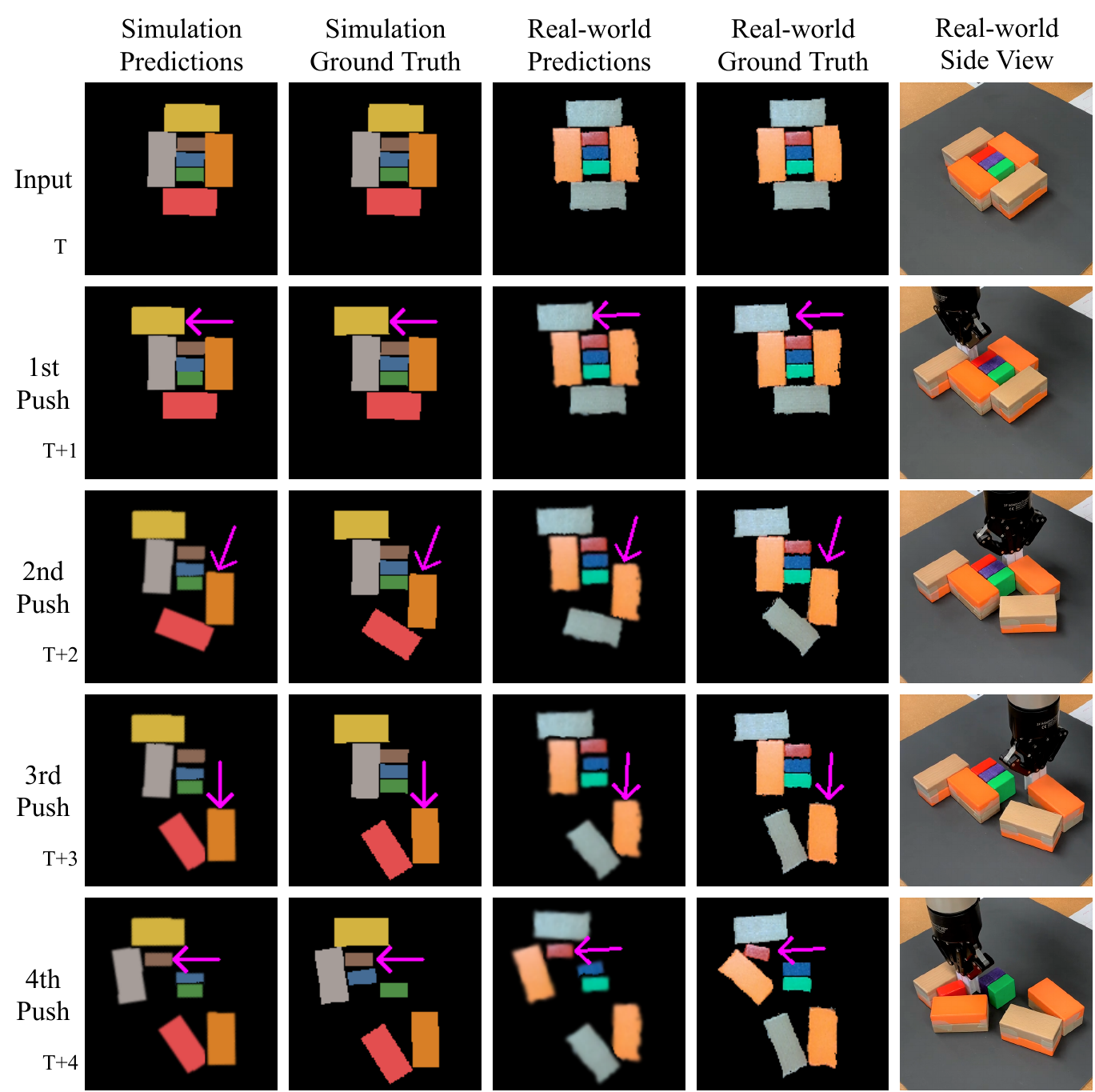}
    \caption{\label{fig:predictions}
        Example of $4$ consecutive pushes showing that 
        \dipn can accurately predict push outcomes over a long horizon. 
        We use purple arrows to illustrate push actions. 
        The first and second columns are the predictions 
        and ground truth (objects' positions after executing the pushes) 
        in simulation. 
        The third and fourth columns show result on a real system. 
        The last column is the side view of the push result.
        Each row represents the push outcome with the previous row 
        as the input observation. 
    }
    \vspace{-2mm}
\end{figure}

The de-cluttering tasks considered in~\cite{huang2020dipn} required only 
single-step predictions. The \orc challenge requires highly accurate predictions 
for multiple consecutive pushes in the future. To adapt \dipn for \orc, we fine-tuned
its architecture, replacing ResNet-18 with ResNet-10~\cite{he2016deep} 
while increasing the dimension of outputs from $256$ to $512$ to predict 
motions of more objects simultaneously and efficiently. The number of decoder MLP layers is 
also increased to six, with sizes $[768, 256, 64, 16, 3, 3]$. Other augmentations 
are reported in Section~\ref{sec:experiments}. Finally, we trained the 
network with $200,000$ random push actions applied on various objects.
This number is higher than the $1,500$ actions used in~\cite{huang2020dipn}
as we aim for the accuracy needed for long-horizon visual foresight. 
Given a sequence of candidate push actions, the fine-tuned DIPN predicts complex interactions, e.g., Fig.~\ref{fig:predictions}.

\subsection{Visual Foresight Tree Search (\ours)}\label{subsec:vft}
We introduce \dipn for predicting single-step push outcome and \gn for 
generating/rating grasps as building blocks for a multi-step procedure capable 
of long-horizon planning. A natural choice is Monte-Carlo Tree Search 
(\mcts)~\cite{mcts2012}, which balances scalability and optimality.
In essence, \ours fuses \mcts and \dipn to generate an optimal multi-step 
push prediction, as graded by \gn. 
A search node in \ours corresponds to an input scene or one imagined by \dipn. 
\mcts prioritizes the most promising states when expanding the search tree; 
in \ours, such states are the ones leading to a successful target retrieval 
in the least number of pushes. 
%
%

In a basic search iteration, \mcts has four essential steps: 
selection, expansion, simulation, and back-propagation. 
First, the {\em selection} stage samples a search node and a push action based on a selection function. 
Then, the {\em expansion} stage creates a child node of the selected node. 
After that, the reward value of the new child node is determined by 
a {\em simulation} from the node to an end state. 
Finally, the {\em back-propagation} stage updates the estimated Q-values of 
the parent nodes.


For describing \mcts with visual foresight, let $N(n)$ be the number of visits 
to a node $n$ and $Q(n) = \{r_1, \dots, r_{N(n)}\}$ as the estimated Q-values of each visit.
We use $N_{max}$ to denote the number of iterations the MCTS performed; 
we may also use an alternative computational budget to stop the search~\cite{mcts2012}.
The high-level workflow of our algorithm is depicted in Alg.~\ref{alg:vft}, and illustrated in Fig.~\ref{fig:pipeline}.
We will describe one iteration (line~\ref{alg:iteration}-\ref{alg:bac-end}) 
of \mcts in \ours along with the pseudo-code in the remaining of this section.


\newcommand\mycommfont[1]{\footnotesize\normalfont\textcolor{gray}{#1}}
\SetCommentSty{mycommfont}
\begin{algorithm}
    \begin{small}
    \DontPrintSemicolon
    \SetKwFunction{FMain}{VFT}
    \SetKwFunction{FMCTS}{MCTS}
    \Fn{\FMain{$s_t$}}{
        \While{\normalfont there is a target object in workspace}{
            $R(s_t) \gets \gn(s_t)$\;
            \If{\normalfont $\max_{a^{\text{grasp}}} R(s_t, a^{\text{grasp}}) > R_{g}^*$}{Execute $\argmax_{a^{\text{grasp}}}R(s_t, a^{\text{grasp}})$\tcp*[f]{Grasp}}
            \lElse{Execute \FMCTS{$s_t$}\tcp*[f]{Push}}
        }
    }
    \vspace*{3pt}
    \SetKwProg{Pn}{Function}{:}{}
    \Pn{\FMCTS{$s_t$}}{
        Create root node $n_0$ with state $s_t$ \;
        $N(\cdot) \gets 0$, $Q(\cdot) \gets \varnothing$\tcp*{Default $N$, $Q$ for a search node}
        \For{$i \gets 1,  2, \dots,  N_{max}$}{
            $n_c \gets n_0$ \;\label{alg:iteration}
            \Comment{\bf Selection and Expansion}
            \While{$n_c$ \normalfont is not expandable}{\label{alg:selection}
                $n_c \gets \pi_{\text{tree}}(n_c)$\label{alg:tree-policy}
                \tcp*{Use (\ref{equation:uct}) to find a child node}
            } 
            \vspace{1mm}
            $a^{\text{push}} \gets$ sample from untried push actions in $n_c$ \label{alg:expansion1}\;
            $n_c \gets \dipn(n_c, a^{\text{push}})$\label{alg:expansion2}\tcp*{Generate node by push prediction}
            \Comment{\bf Simulation}
            $r \gets 0$, $d \gets 1$, $s \gets n_c.\text{state}$\label{alg:sim-start}\tcp*{$s$ is the state of $n_c$}
            \While{$s$ \normalfont is not a terminal state}{
                $a^{\text{push}} \gets$ randomly select a push action in $s$\label{alg:simulation-roll} \;
                $s \gets \dipn(s, a^{\text{push}})$ \label{alg:simulation-pred}\tcp*{Simulate to next state}
                $R(s) \gets \gn(s)$\;
                $r \gets \max\{r, \gamma^{d} \max_{a^{\text{grasp}}} R(s, a^{\text{grasp}})\}$\; \label{alg:sim-gn}
                $d \gets d + 1$
            } \label{alg:sim-end}
            \Comment{\bf Back-propagation}
            \While{$n_c$ \normalfont is not root\label{alg:bac-start}}{
                $N(n_c) \gets N(n_c) + 1$ \;
                $R(n_c\text{.state}) \gets \gn(n_c\text{.state})$\;
                $r \gets \max\{r, \max_{a^{\text{grasp}}} R(n_c\text{.state}, a^{\text{grasp}})\}$ \; \label{alg:bac-gn}
                $Q(n_c) \gets Q(n_c) \cup \{r\}$\tcp*{Record the reward}
                $r \gets r \cdot \gamma$\; 
                $n_c \gets \text{parent of } n_c$
            } \label{alg:bac-end}
        }
        $n_{\text{best}} \gets \argmax_{n_i \in \text{children of } n_0}(\uct(n_i, n_0))$ \label{alg:best-action} \;
        \Return push action $ a^{\text{push}}$ that leads to $n_{\text{best}}$ from the root
    }
    \caption{\label{alg:vft}
    Visual Foresight Tree Search}
    \end{small}
\end{algorithm}


\textbf{Selection.} 
The first step of \mcts is to select an {\em expandable} search node
(line~\ref{alg:selection}-\ref{alg:tree-policy}) using a tree policy 
$\pi_\text{tree}$. 
Here, {\em expandable} means the node has some push actions that 
are not tried via selection-expansion; more details of the push action space 
will be discussed later in the expansion part. 
To balance between exploration and exploitation, 
when the current node $n_c$ is already fully expanded, 
$\pi_\text{tree}$ uses Upper Confidence Bounds for Trees (\uct)~\cite{mcts2012} 
to rank its child node $n_i$. We customize \uct as
\begin{equation}\label{equation:uct}
    \uct(n_i, n_c) = \frac{Q^{m}(n_i)}{\min\{N(n_i), m\}} + 
    C \sqrt{\frac{\ln{N(n_c)}}{N(n_i)}}.  
\end{equation}

Here, $C$ is an exploration weight. 
In the first term of (\ref{equation:uct}), unlike  typical \uct that favours 
the child node that maximizes $Q(n_i)$, 
we keep only the most promising rollouts of $n_i$ and denote by $Q^m(n_i)$ the average returns of the top $m$ rollouts of $n_i$. 
In our implementation, $m = 3$ and $C = 2$. 
We also use (\ref{equation:uct}) with parameters $m = 1$ and $C = 0$ 
to find the best node, and thus the best push action to execute, after the search is completed, as shown in line~\ref{alg:best-action}.

\textbf{Expansion.} 
Given a selected node $n$, we use \dipn to generate a child node 
by randomly choosing an untried push action $a^\text{push}$
(line~\ref{alg:expansion1}-\ref{alg:expansion2}). 
The action $a^\text{push}$ is uniformly sampled at random from the 
selected node's action space, 
which contains two types of push actions:
\begin{enumerate*}
    \item 
    For each object, we apply principal component analysis to 
    compute its feature axis. 
    For example, for a rectangle object, the feature axis will be parallel to 
    its long side. 
    Four push actions are then sampled with directions perpendicular 
    or parallel to the feature axis, 
    pushing the object from the outside to its center.
    \item 
    To build a more complete action space, eight additional actions are evenly 
    distributed on each object's contour, with push direction also 
    towards the object's center.  
\end{enumerate*}

\textbf{Simulation.}
After we generated a new node via expansion, 
in line~\ref{alg:sim-start}-\ref{alg:sim-end}, 
we estimate the node's Q-value by 
uniformly randomly select push actions at random (line~\ref{alg:simulation-roll}) 
and use \dipn to predict future states (line~\ref{alg:simulation-pred}) 
until one of the following two termination criteria is met:
\begin{enumerate*}
    \item The total number of push actions used to reach a simulated state 
    is larger than a constant $D^*$.
    \item 
    The maximum predicted reward value of a simulated state 
    exceeds a threshold $R_{gp}^*$.
\end{enumerate*}
In line~\ref{alg:sim-gn}, when calculating $r$, a discount factor $\gamma$ 
is used to penalize a long sequence of action. 
Here, we use $\max\gn$ to reference the maximum value in a grasp reward table.  
In our implementation, \gn is only called once for each unique state 
and the output is saved by a hashmap.

\textbf{Back-propagation.} \label{back-propagation}
After simulation, the terminal grasp reward is back-propagated 
(line~\ref{alg:bac-start}-\ref{alg:bac-end})
through its parent nodes to update their $N(n)$ and $Q(n)$.
Denote by $r_0$ the max grasp reward of a newly expanded 
node $n_0$, and $n_1, n_2, \dots, n_k$ as the sequence of $n_0$'s parents in the ascending order up to node $n_k$. With $Q(n_0) = \{r_0\}$, the Q-value of $n_k$ in this iteration 
is then $\max_{0 \leq j < k} \gamma^{k - j}\max Q(n_j)$, 
which corresponds to the max reward of states 
along the path~\cite{DBLP:journals/corr/abs-1912-07024}.
Here, $\gamma$ is a discount factor to penalize a long sequence of actions. 
As a result, for each parent $n_k$, $N(n_k)$ increases by $1$, and 
$\max_{0 \leq j < k} \gamma^{k - j}\max Q(n_j)$ 
is added to $Q(n_k)$. 

\section{Experimental Evaluation}\label{sec:experiments}
We performed an extensive evaluation of the proposed method, \ours, in simulation 
and on the real hardware system illustrated in Fig.~\ref{fig:intro}. \ours is compared with multiple state-of-the-art
approaches~\cite{zeng2018learning, huang2020dipn, xu2021efficient}, with 
necessary modifications for solving \orc, i.e., minimizing the number of actions
in retrieving a target. 
The results convincingly demonstrate \ours to be robust and more efficient 
than the compared approaches. 
%

Both training and inference are performed on a machine with an Nvidia GeForce 
RTX 2080 Ti graphics card, an Intel i7-9700K CPU, and 32GB of memory.


\begin{figure}[ht!]
\vspace{-1mm}
    \centering
    \includegraphics[width = \linewidth]{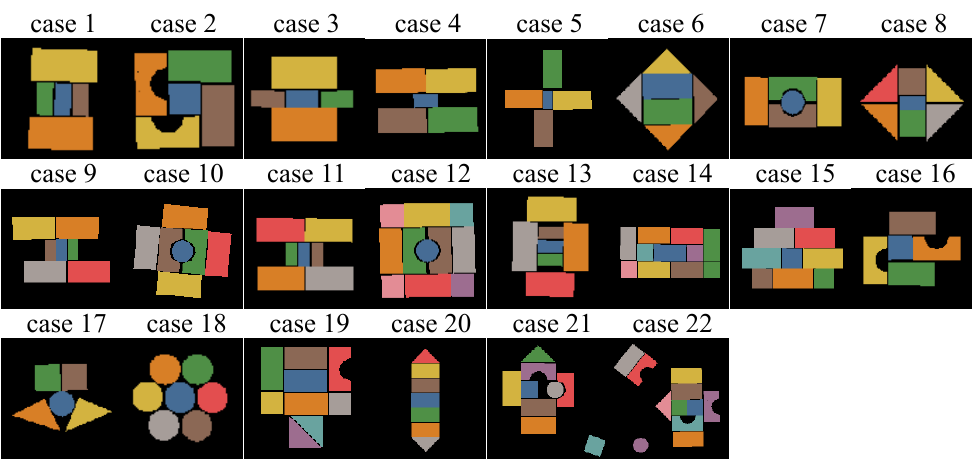}
    \caption{\label{fig:testcases}
        22 Test cases used in both simulation and real world experiments. 
        The target objects are blue. 
        Images are zoomed in for better visualization.
    }
\vspace{-3mm}    
\end{figure}

\subsection{Experiment Setup}
The complete test case set includes 
\begin{enumerate*}
    \item the full set of $14$ test cases from \cite{xu2021efficient}, and 
    \item $18$ hand-designed and more challenging test cases where the objects are tightly packed. 
\end{enumerate*}
All test cases are constructed using wood blocks with different shapes, colors, 
and sizes. 
We set the workspace's dimensions to $44.8 \text{cm} \times 44.8 \text{cm}$. 
The size of the images is $224 \times 224$. Push 
actions have a minimum $5$cm {\em effective push distance}, 
defined as the end-effector's moving distance after object contact. 
Multiple planned push actions may be concatenated if they are in the 
same direction and each action's end location is the same as the next 
action's start location. 
In all scenes, the target object is roughly at the center of the scene.

The hyperparameters for \ours are set as follows. The number of iterations 
$N_{max}= 150$. The discount factor $\gamma=0.8$. The maximum depth $D^*$ of the tree is capped at $4$. The terminal threshold of grasp reward $R_{gp}^* = 1.0$. Threshold $R_g^*$ that decides to grasp or to push is $0.8$ in the simulation experiments and $0.7$ in the real hardware experiments. 
Such thresholds can potentially be fully optimized for a production system; it is not carried out in this work as reasonably good values are easily obtained while it is prohibitively time-consuming to carry out a full-scale optimization.




\subsection{Network Training Process}
\ours contains two deep neural networks: \gn and \dipn. 
Both are trained in simulation with the same objects as used in real experiments to capture the physical properties and dynamics of the environment. 
No prior knowledge is given to the networks except the dimensions of the gripper fingers.

\gn is trained on-policy with \num{20000} grasp actions. Similar to 
\cite{zeng2018learning, huang2020dipn, xu2021efficient}, randomly-shaped objects are uniformly dropped onto the workspace to construct the training scenarios. 
A successful grasp is decided by checking the distance between grippers, which should be greater than 0. 
A Huber loss on the pixel where the robot performed the grasp action is used. 
All other pixels do not contribute to the loss during back-propagation.
Image-based pre-training~\cite{huang2020dipn, yen2020learning} was employed to initialize the training parameters.
We then train the \gn by stochastic gradient descent with the momentum of $0.9$, weight decay of $10^{-4}$, and batch size of $12$.
The learning rate is set to $5 \times 10^{-5}$ and by half every $2000$ iteration.

\dipn \cite{huang2020dipn} is trained in a supervised manner with \num{200000} 
random push actions from simulation. In the push data set, $20\%$ of the scenes contain randomly placed objects, and $80\%$ contain densely packed objects. 
The push distance for DIPN is fixed to 7.4 cm (effective touch distance is 5 cm). 
In the original DIPN paper~\cite{huang2020dipn}, the distance was 5 cm and 10 cm without considering the effective range.

We note that a total of $2000$ actions ($500$ grasps and $1500$ pushes) 
are sufficient for the networks to achieve fairly accurate results
(see, e.g.,~\cite{huang2020dipn}). Because training samples are readily 
available from simulation, it is not necessary to skimp on training data.
We thus opted to train with more data to evaluate the full potential of \ours. 

A Smooth L1 Loss with beta equals to $2$ is used instead of $1$~\cite{huang2020dipn}.
We train the \dipn by stochastic gradient descent with the momentum of $0.9$, weight decay of $10^{-4}$, and using cosine annealing schedule~\cite{loshchilov2016sgdr} with learning rates of learning rate of $10^{-3}$ for $76$ epochs, and the batch size is $128$.

\subsection{Compared Methods and Evaluation Metrics}
\textbf{Goal-Conditioned VPG (\gcvpg).} Goal-conditioned VPG (\gcvpg) is a modified 
version of Visual Pushing Grasping (VPG)~\cite{zeng2018learning}, which uses two 
DQNs \cite{mnih2015human} for pushing and grasping predictions. VPG by itself does not 
focus on specific objects; it was conditioned \cite{xu2021efficient} to focus on the 
target object to serve as a comparison point, yielding \gcvpg. 

\textbf{Goal-Oriented Push-Grasping.} In 
\cite{xu2021efficient}, many modifications are applied to VPG to render the resulting 
network more suitable for solving \orc, including adopting a three-stage training 
strategy and an efficient labeling method \cite{NIPS2017_453fadbd}. For convenience, 
we refer to this method as \gopg (the authors of \cite{xu2021efficient} did not provide a 
short name for the method).

\textbf{\dipn.}
As an ablation baseline for evaluating the utility of employing deep tree search, we replace 
\mcts from \ours with a search tree of depth one. In this baseline, \dipn is  used to evaluate 
all candidate push actions. The push action whose predicted next state has the 
highest grasp reward for the target object is then chosen. This is similar to 
how \dipn is used in \cite{huang2020dipn}; we thus refer to it simply as \dipn.

In our evaluation, the main metric is the total number of push and grasp 
actions used to retrieve the target object. 
For a complete comparison to \cite{zeng2018learning, xu2021efficient}, 
we also list \ours's grasp success rate, which is the ratio of successful grasps 
in the total number of grasps during testing. The completion rate, i.e., the chance of eventually grasping the target object, is also reported. Similar to
\cite{huang2020dipn}, when \dipn is used, a $100\%$ completion rate often reached.

We only collected evaluation data on \dipn and \ours. For the other two baselines, 
\gcvpg and \gopg, results are directly quoted from \cite{xu2021efficient} (at 
the time of our submission, we could not obtain the trained model or the information
necessary for the reproduction of \gcvpg and \gopg). While our hardware setup is identical to that of~\cite{xu2021efficient}, and the poses of objects are also identical,
we note that there are some small
differences between the evaluation setups:
\begin{enumerate*}
    \item We use PyBullet~\cite{coumans2019} for simulation, 
    while \cite{xu2021efficient} uses CoppeliaSim~\cite{6696520}; 
    the physics engine is the same (Bullet). 
    \item \cite{xu2021efficient} uses an RD2 gripper in simulation and 
    a Robotiq 2F-85 gripper for real experiment; 
    all of our experiments use 2F-85. 
    \item \cite{xu2021efficient} has a $13$cm push distance, 
    while we only use a $5$cm effective distance 
    (the distance where fingers touch the objects)
    \item \cite{xu2021efficient} uses extra top-sliding pushes which 
    expand the push action set. 
\end{enumerate*}
At the same time, we confirm that these relatively minor differences do not 
provide our algorithm any unfair advantage.

\subsection{Simulation Studies}
\begin{figure}[ht!]
    \centering
    \includegraphics[width = .97\linewidth]{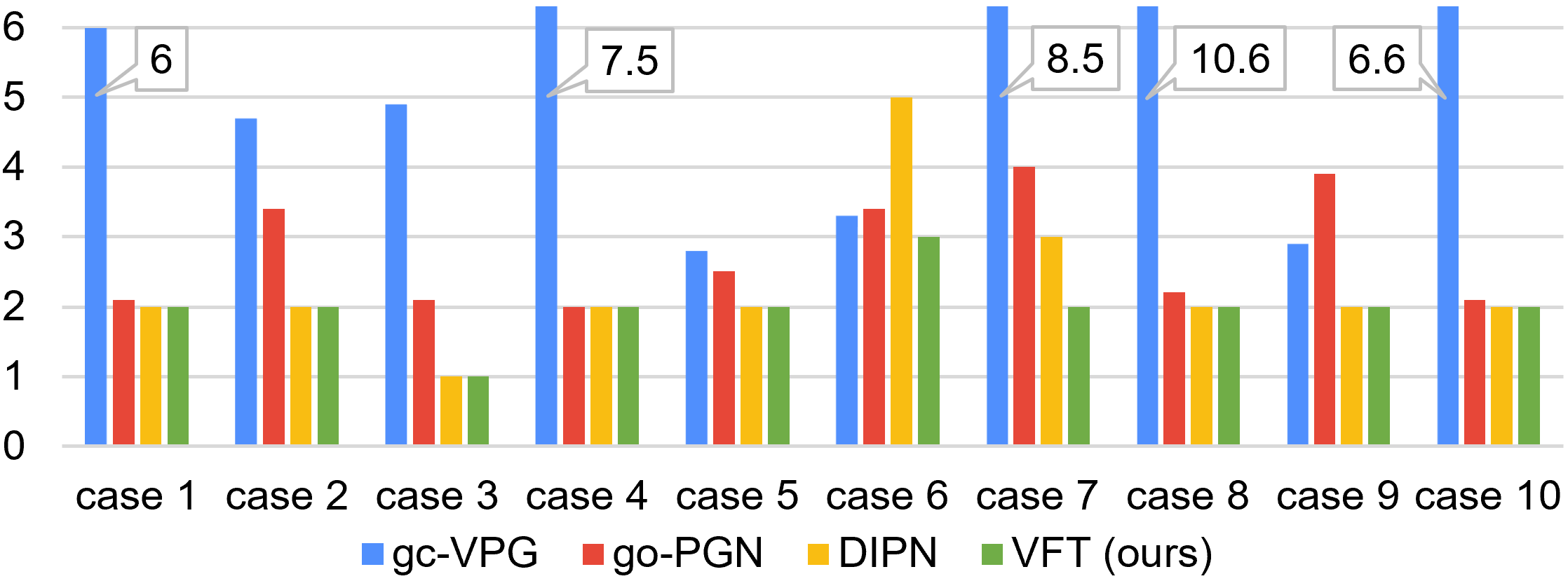}
    \caption{\label{fig:baseline-hist}
        Simulation results per test case for the $10$ problems from 
        \cite{xu2021efficient}.
        The horizontal axis shows the average number of actions used to solve 
        a problem instance: the lower, the better. 
    } 
\vspace{-2mm}    
\end{figure}

Fig.~\ref{fig:baseline-hist} and Table.~\ref{tab:10table} show the evaluation results 
of all algorithms on the $10$ simulation test cases from \cite{xu2021efficient}. Each experiment is   repeated $30$ times, and the average number of actions until task completion in each experiment is reported.
%
Our proposed method, \ours, which uses an average of $2.00$ actions, significantly 
outperforms the compared methods. Specifically, \ours uses one push action and one 
grasp action to solve the majority of cases, except for one instance with a 
half-cylinder shaped object, which is not included during the training of the 
networks. Interestingly, when only one push is necessary, \ours, with its main 
advantage as multi-step prediction, still outperforms \dipn due to its extra 
simulation steps. 
The algorithms with push prediction performs better  than \gcvpg and \gopg in all metrics.

\begin{table}[ht!]
\vspace{-1mm}    
    \centering
    \begin{tabular}{c|c|c|c}
        & Completion & Grasp Success & Number of Actions \\ \hline
        \gcvpg \cite{xu2021efficient} & $89.3\%$ & $41.7\%$ & $5.78$ \\ \hline
        \gopg \cite{xu2021efficient} & $99.0\%$ & $90.2\%$ & $2.77$ \\ \hline
        \dipn \cite{huang2020dipn}  & $100\%$ & $100\%$ & $2.30$ \\ \hline
        \ours (ours) & $100\%$& $100\%$ & $\mathbf{2.00}$ \\ \hline
    \end{tabular}
    \caption{Simulation results for the $10$ test cases from \cite{xu2021efficient}.}
    \label{tab:10table}
\vspace{-2mm}
\end{table}

\begin{figure}[ht!]
    \centering
    \includegraphics[width = .97\linewidth]{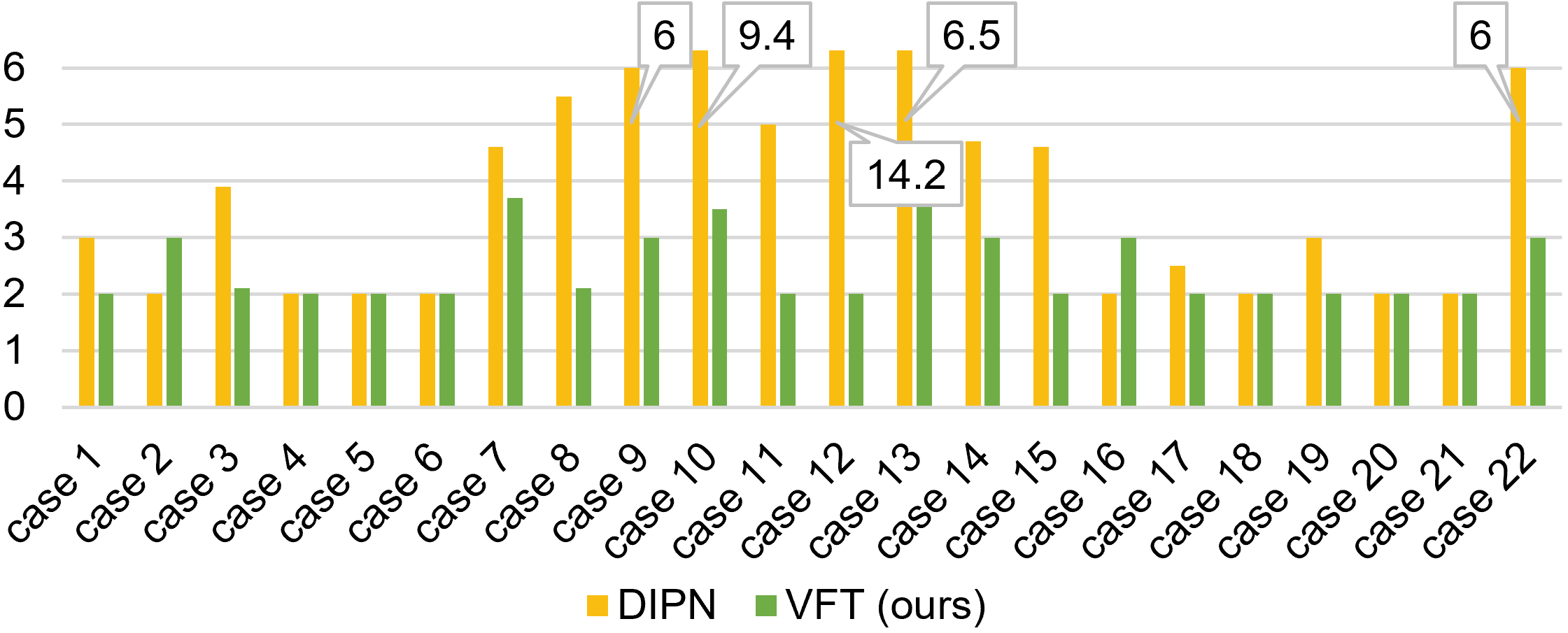}
    \caption{\label{fig:bar-sim-22}
        Simulation result per test case for the $22$ harder problems 
        (Fig.~\ref{fig:testcases}).
        The horizontal axis shows the average number of actions used to solve 
        a problem instance: the lower, the better.
    }
\end{figure}

To probe the limit of \ours's capability, we evaluated the methods on harder cases 
demanding multiple pushes. The test set includes $18$ manually designed instances 
and $4$ cases from \cite{xu2021efficient} (see Fig.~\ref{fig:testcases}). 
As shown in Fig.~\ref{fig:bar-sim-22} and Table.~\ref{tab:22table-sim}, \ours uses 
fewer actions than \dipn as \ours looks further into the future. Though we could not 
evaluate the performance of \gcvpg and \gopg on these settings for direct comparison because we could not obtain the information necessary for the reproduction of these systems,
notably, the average number of actions ($2.45$) used by \ours on harder instances is 
even smaller than the number of actions ($2.77$) \gopg used on the $10$ simpler cases. 

\begin{table}[ht!]
\vspace{-1mm}
    \centering
    \begin{tabular}{c|c|c|c}
        & Completion & Grasp Success & Num. of Actions \\ \hline
        \dipn \cite{huang2020dipn} & $100\%$ & $98.3\%$ & $4.31$ \\ \hline
        \ours (ours) & $100\%$ & $98.8\%$ & $\mathbf{2.45}$ \\ \hline
        
    \end{tabular}
    \caption{Simulation result for the $22$ test cases in 
    Fig.~\ref{fig:testcases}.}
    \label{tab:22table-sim}
\vspace{-3mm}    
\end{table}

\subsection{Evaluation on a Real System}

\begin{figure}[ht!]
    \centering
    \includegraphics[width = .97\linewidth]{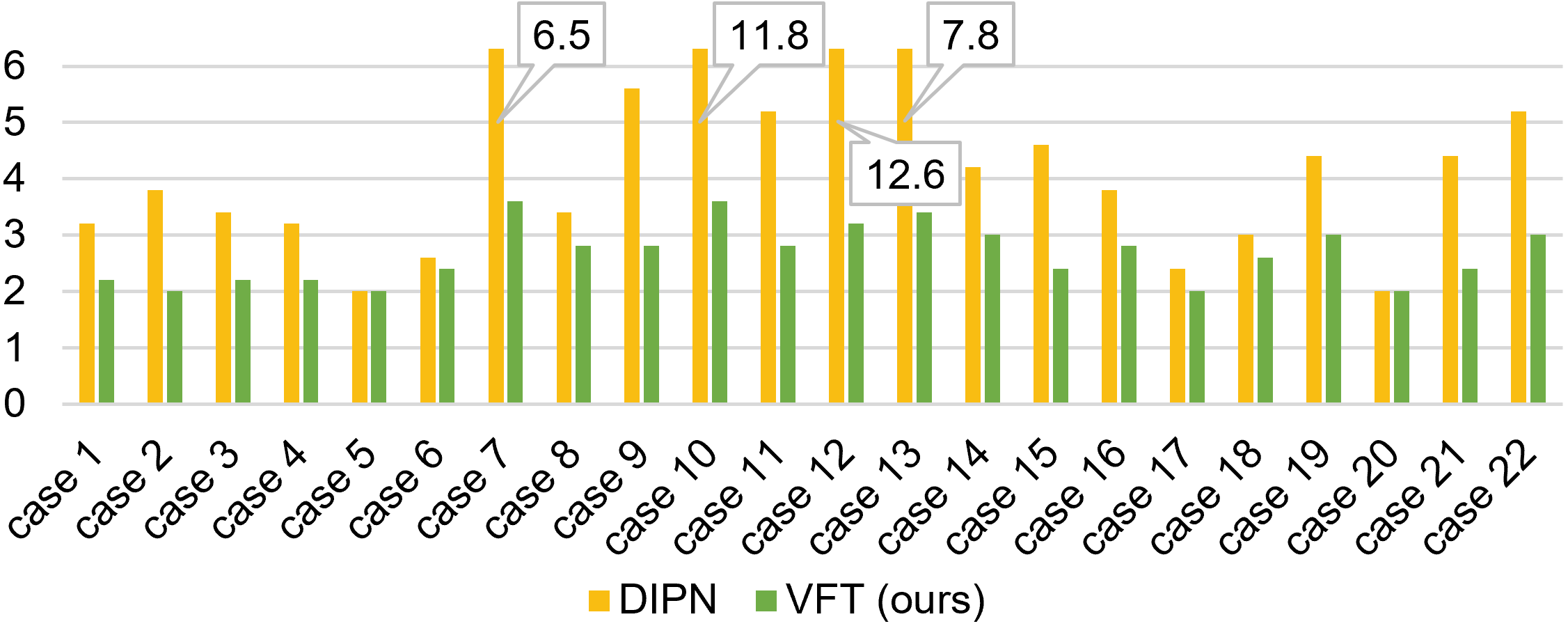}
    \caption{\label{fig:bar-real-22}
        Real experiment results per test case for the $22$ harder problems 
        (Fig.~\ref{fig:testcases}).
        The horizontal axis shows the average number of actions used to solve 
        a problem instance: the lower, the better.
    }
\vspace{-3mm}    
\end{figure}

We repeated the $22$ hard test cases on a real robot system (Fig.~\ref{fig:intro-setup}). 
Both \ours and \dipn are evaluated. We also bring the experiment result from
\cite{xu2021efficient} on its $4$ real test cases for comparison. All cases are 
repeated at least 5 times to get the mean metrics. The result, shown in Fig.~\ref{fig:bar-real-22}, Table.~\ref{tab:22table-real}, and Table.~\ref{tab:4table} 
closely matches the results from simulation. We observe a slightly lower grasp 
success rate due to the more noisy depth image on the real system. The real 
workspace's surface friction is also different from simulation. However, \ours and 
\dipn can still generate accurate foresight. 

\begin{table}[ht!]
    \centering
    \begin{tabular}{c|c|c|c}
        & Completion & Grasp Success & Num. of Actions \\ \hline
        \dipn \cite{huang2020dipn} & $100\%$ & $97.0\%$ & $4.78$ \\ \hline
        \ours (ours) & $100\%$ & $98.5\%$ & $\mathbf{2.65}$ \\ \hline
    \end{tabular}
    \caption{Real experiment results for the $22$ Test cases in 
    Fig.~\ref{fig:testcases}.}
    \label{tab:22table-real}
\vspace{-2mm}    
\end{table}

\begin{table}[ht!]
    \centering
    \begin{tabular}{c|c|c|c}
        & Completion & Grasp Success & Num. of Actions \\ \hline
        \gopg \cite{xu2021efficient} & $95.0\%$ & $86.6\%$ & $4.62$ \\ \hline
        \dipn \cite{huang2020dipn} & $100\%$ & $100\%$ & $4.00$ \\ \hline
        \ours (ours) & $100\%$ & $100\%$ & $\mathbf{2.60}$ \\ \hline
        
    \end{tabular}
    \caption{Real experiment results for cases $19$ to $22$ in 
    Fig.~\ref{fig:testcases}.}
    \label{tab:4table}
\vspace{-2mm}    
\end{table}

\begin{figure}[ht!]
    \centering
    \includegraphics[width=0.24\linewidth]{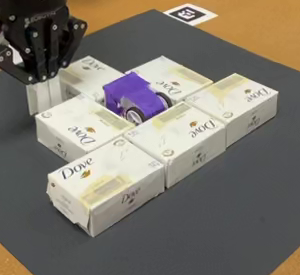}
    \hfill
    \includegraphics[width=0.24\linewidth]{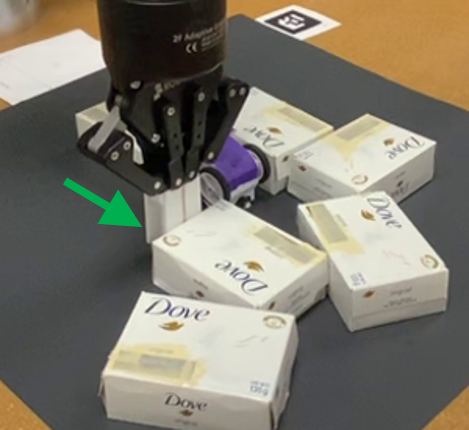}
    \hfill
    \includegraphics[width=0.24\linewidth]{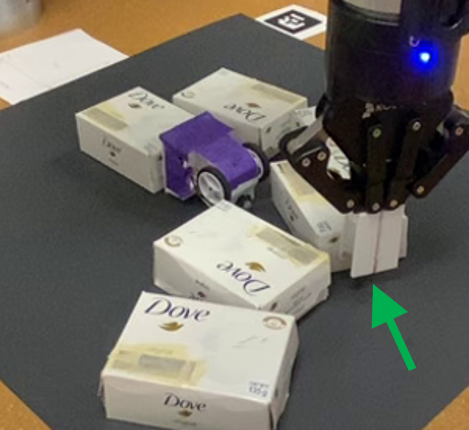}
    \hfill
    \includegraphics[width=0.24\linewidth]{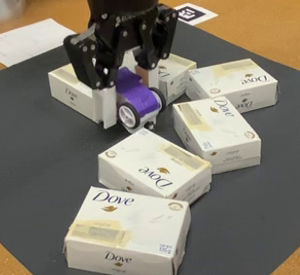}
    \caption{\label{fig:car-result} 
        Test scenario with soap boxes and masked 3D printed vehicle. 
        Two push actions and one grasp action.
    }
\vspace{-2mm}
\end{figure}

We also explored our system on everyday objects (Fig.~\ref{fig:car-result}),  
where we want to retrieve a small robotic vehicle surrounded by soapboxes. 
Although the soapboxes and the small vehicles are unseen types of objects during training, the robot is 
able to strategically push the soapboxes away in two moves only and retrieve the vehicle. 

We report that the running time to decide one push action is around 
$3$ minutes on average when the number of \mcts iterations is set to be $150$.
A single push prediction of \dipn took $30$ milliseconds.
While using the simulator as the transition function in \mcts under a similar criterion would take $8$ minutes on average to decide one push action.
In this letter, our primary focus is action optimization. 


\section{Conclusion and Discussions}\label{sec:conclusion}
In conclusion, through an organic fusion of Deep Interaction Prediction Network (\dipn) and MCTS, the proposed Visual Foresight Trees (\ours) can make a high-quality multi-horizon prediction for optimized object retrieval from dense clutter. The effectiveness of \ours is convincingly demonstrated with extensive evaluation.
As to the limitations of \ours, the time required is relatively long because of the large \mcts tree that needs to be computed. This can be improved with multi-threading because the rollouts have sufficient independence. 
Currently, only a single thread is used to complete the \mcts.
It would also be interesting to develop a network for directly estimating the reward for rollout policy, which would reduce the inference time.
This technique would be similar in spirit to the MuZero algorithm~\cite{schrittwieser2020mastering}, which has been shown to be efficient by combining Monte Carlo tree search and learning by self-playing in an end-to-end manner. The learned rollout policy could lead to better performance. One issue related to end-to-end training is data efficiency, which is why this type of technique has been limited to games. Improving the data efficiency of end-to-end techniques is crucial to the deployment of these techniques on robotic tasks.



%





\ifCLASSOPTIONcaptionsoff
  \newpage
\fi



%




\def\url#1{}
\bibliographystyle{formatting/IEEEtran}
\bibliography{bib}

%








\end{document}